%% file: root.tex
\documentclass[letterpaper, 10 pt, conference]{ieeeconf}  

\IEEEoverridecommandlockouts                              

\usepackage{graphics} 
\usepackage{epsfig} 
\usepackage{times} 
\usepackage{amsmath} 
\usepackage{amssymb}  
\usepackage{authblk}
\usepackage{booktabs}
\usepackage{makecell}
\DeclareMathAlphabet{\mathcal}{OMS}{cmsy}{m}{n}
\DeclareSymbolFont{largesymbols}{OMX}{cmex}{m}{n}
\usepackage{color}
\usepackage{hyperref}

\title{\LARGE \bf
Advancing Dense Endoscopic Reconstruction with Gaussian Splatting-driven Surface Normal-aware Tracking and Mapping
}

\author{Yiming Huang$^{1,2}$*, Beilei Cui$^{1,2}$*, Long Bai$^{1,2}$*, Zhen Chen$^{3}$, Jinlin Wu$^{3}$, Zhen Li$^{4}$, \\ Hongbin Liu$^{3}$, \textit{Senior Member, IEEE}, and Hongliang Ren$^{1,2,\dagger}$, \textit{Senior Member, IEEE}
\thanks{$^\dagger$This work was supported by Hong Kong RGC CRF C4026-21GF, GRF 14203323, GRF 14216022, GRF 14211420, NSFC/RGC Joint Research Scheme N\_CUHK420/22, Shenzhen-Hong Kong-Macau Technology Research Programme (Type C) STIC Grant 202108233000303. (Corresponding author: H. Ren.)}
\thanks{$^{1,2}$Y. Huang, B. Cui, L. Bai (*: Equally contribution), and H. Ren are with the Dept. of Electronic Engineering, The Chinese University of Hong Kong (CUHK), Hong Kong, China; and also with the CUHK Shenzhen Research Institute, Shenzhen, China. (E-mail: \{yhuangdl, beileicui, b.long\}@link.cuhk.edu.hk)}
\thanks{$^{3}$J. Wu, Z. Chen, and H. Liu are with the Centre for Artificial Intelligence and Robotics (CAIR), Hong Kong Institute of Science \& Innovation, Chinese Academy of Sciences, Hong Kong, China.}
\thanks{$^{4}$Z. Li is with Qilu Hospital of Shandong University, Jinan, China.}
}

\begin{document}

\maketitle
\thispagestyle{empty}
\pagestyle{empty}
\input{sections/0_abstract}
\input{sections/1_introduction}
\input{sections/2_related_works}
\input{sections/3_system}

\input{sections/4_experiment}
\input{sections/5_conclusions}


\bibliographystyle{IEEEtran}
\bibliography{references}
\end{document}

%% file: sections/0_abstract.tex
\begin{abstract}
Simultaneous Localization and Mapping (SLAM) is essential for precise surgical interventions and robotic tasks in minimally invasive procedures. While recent advancements in 3D Gaussian Splatting (3DGS) have improved SLAM with high-quality novel view synthesis and fast rendering, these systems struggle with accurate depth and surface reconstruction due to multi-view inconsistencies. Simply incorporating SLAM and 3DGS leads to mismatches between the reconstructed frames. In this work, we present Endo-2DTAM, a real-time endoscopic SLAM system with 2D Gaussian Splatting (2DGS) to address these challenges. Endo-2DTAM incorporates a surface normal-aware pipeline, which consists of tracking, mapping, and bundle adjustment modules for geometrically accurate reconstruction. Our robust tracking module combines point-to-point and point-to-plane distance metrics, while the mapping module utilizes normal consistency and depth distortion to enhance surface reconstruction quality. We also introduce a pose-consistent strategy for efficient and geometrically coherent keyframe sampling. Extensive experiments on public endoscopic datasets demonstrate that Endo-2DTAM achieves an RMSE of $1.87\pm 0.63$ mm for depth reconstruction of surgical scenes while maintaining computationally efficient tracking, high-quality visual appearance, and real-time rendering. Our code will be released at~\href{https://github.com/lastbasket/Endo-2DTAM}{github.com/lastbasket/Endo-2DTAM}.

\end{abstract}

%% file: sections/1_introduction.tex
\section{Introduction}
Endoscopic surgery~\cite{de2017history} has revolutionized the medical field by enabling minimally invasive procedures that reduce patient recovery time and scarring. This technique allows surgeons to access internal organs through small incisions, reducing the need for large-scale surgical interventions. Despite these benefits, the intricate and confined spaces within the human body present a formidable challenge for surgeons~\cite{eliashar2003image}. The narrow visual field and lack of depth perception typical of endoscopic imaging can impede the surgeon's ability~\cite{marcus2014endoscopic} to interpret and navigate the complex 3D environment accurately. To mitigate the risks and enhance surgical outcomes, there is a growing need for visualization tools~\cite{fu2021future, chen2023surgical} that can provide real-time, high-fidelity 3D reconstructions of surgical scenes.

\begin{figure}[t!]
    \label{fig1}
    \centering
    \includegraphics[width=0.94\linewidth]{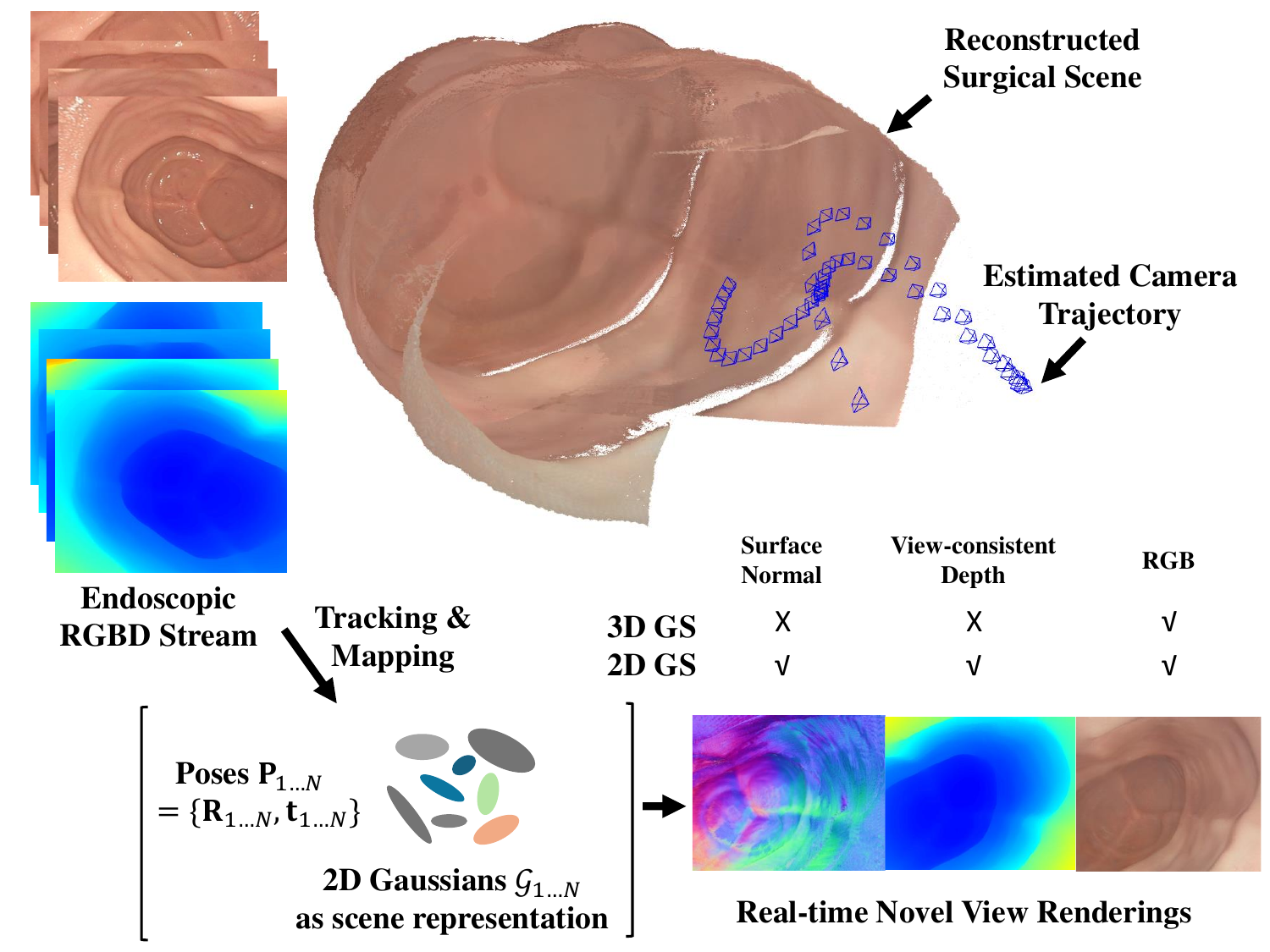}
    \caption{\textbf{Reconstruction and Rendering Results.} Compared with the 3DGS-based SLAM, our method utilizes 2DGS for geometry-accurate scene representation, producing novel view rendering of high-quality images, view-consistent depth maps, and precise surface normal.
}
\end{figure}

To address these challenges, Simultaneous Localization and Mapping (SLAM) techniques~\cite{ali2022we, azagra2023endomapper} have been widely used for estimating the camera pose and reconstructing the surgical scene in real time. Traditional SLAM systems~\cite{grasa2013visual, mahmoud2017slam, wang2019visual} demonstrate precise camera tracking and efficient runtime but lack dense geometry and texture information~\cite{kazerouni2022survey}. \textit{e.g.} the widely used ORB-SLAM3~\cite{orbslam} has made significant strides in precise camera tracking and efficient runtime performance but relies on post-process volumetric fusion for reconstruction. While recent methods~\cite{gu2022vision, sageslam, ma2021rnnslam, posner2023c, rau2023bimodal, endodepth, afsfm, wei2022stereo} provide depth map estimation for dense reconstruction, they still struggle with fine-grained dense reconstructions. The recent advances in neural rendering by using Neural Radiance Fields (NeRF)~\cite{nerf} has provided novel view synthesis for dense surgical scene reconstruction~\cite{endonerf, forplane, lerplane, huang2024endo4dgsendoscopicmonocularscene, liu2024endogaussian, zhu2024deformable}. The neural implicit field facilitates SLAM systems~\cite{li2023dense, imap, wang2023co, niceslam, zhu2023nicer, sandstrom2023point} with high-quality novel view image and depth maps. For example, the pioneering works iMap~\cite{imap} and NICE-SLAM~\cite{niceslam} utilize the multi-layer perceptron (MLP) and neural implicit grid for scene representations. However, their ray-based volume rendering requires high computational costs and fails to balance the trade-off between efficiency and accuracy.  On the other hand, 3D Gaussian Splatting (3DGS)~\cite{3dgs}-based SLAM systems~\cite{Matsuki:Murai:etal:CVPR2024, Yan_2024_CVPR, keetha2024splatam}, deliver both high-speed and high-fidelity reconstructions for novel view rendering, offering new possibilities for real-time visualization and mapping within surgical applications~\cite{wang2024endogslam}.

Despite the advantages, 3DGS~\cite{3dgs} is unable to render surface normal and fails to reconstruct accurate depth and surface detail due to the multi-view inconsistencies~\cite{Huang2DGS2024}. Such limitations are particularly problematic in surgical contexts, where an exact spatial representation is crucial. To address these challenges, we present Endo-2DTAM, an innovative real-time tracking and mapping system with 2D Gaussian Splatting (2DGS)~\cite{Huang2DGS2024} for endoscopic reconstruction, as seen in Fig.~\ref{fig1}. Our proposed system encompasses three modules: the tracking module, the mapping module, and bundle adjustment. While 2DGS facilitates explicit representation for surface normal~\cite{Huang2DGS2024}, our tracking module utilizes the point-to-point and point-to-plane distances simultaneously, facilitating localization with surface normal information. Subsequently, our mapping module jointly accounts for normal consistency and depth distortion to enhance the quality of surface reconstruction. Furthermore, we introduce a pose-consistent strategy for keyframe sampling. By optimizing the selected keyframes in mapping and bundle adjustment, we strike a balance between comprehensive coverage and computational efficiency. We are the first to incorporate 2DGS into the SLAM system to address multi-view inconsistency issues that have hindered the prior 3DGS-based SLAM~\cite{wang2024endogslam}. We evaluate our method on the public dataset and achieve state-of-the-art performance for geometry reconstruction, validating the potential for surgical applications.
Overall, our main contributions include the following:
\begin{itemize}
    \item We present a novel SLAM system with 2D Gaussian Splatting for precise camera tracking and high-fidelity 3D tissue reconstruction in endoscopic scenes. 
    \item Our system provides real-time novel view rendering for photorealistic RGB images, view-consistent depth, and accurate surface normal.
    \item We propose a surface normal-aware tracking and mapping pipeline with a pose-consistent keyframing strategy for accurate geometry reconstruction.
    \item We conduct extensive experiments on the public dataset and achieve state-of-the-art geometry reconstruction with only $1.87\pm 0.63$ RMSE (mm) for the depth in endoscopic scene.
\end{itemize}

%% file: sections/2_related_works.tex
\section{Related Work}

\subsection{Neural Field based SLAM}


The recent neural field methods achieve dense reconstruction and high-quality novel view synthesis~\cite{nerf, zhang2020nerf++, martin2021nerf}, which is also extended to endoscopic reconstructions~\cite{endonerf, forplane, lerplane}. The implicit-based methods like iMAP~\cite{imap}, MeSLAM~\cite{kruzhkov2022meslam}, and Go-SLAM~\cite{zhang2023go} utilize MLP for scene representation and achieve satisfactory results. While for explicit field representation, ESLAM~\cite{johari2023eslam}, Point-SLAM~\cite{sandstrom2023point}, and Loopy-SLAM~\cite{liso2024loopy} apply tri-planes and neural point cloud, improving the mapping fidelity. The hybrid-based method such as NICE-SLAM~\cite{niceslam}, NGEL-SLAM~\cite{mao2024ngel}, Co-SLAM~\cite{wang2023co} uses neural grid representation, optimizing the rendering with joint coordinate-based feature and parametric encoding. However, due to the high computational cost of volume rendering, the neural field-based SLAM systems struggle with imbalanced accuracy and efficiency.

\subsection{Gaussian Splatting}

The recent neural rendering using 3D Gaussian splitting~\cite{3dgs} has gained popularity with supreme performance for novel view synthesis in real-time. The later works~\cite{luiten2023dynamic, Wu_2024_CVPR, yang2023gs4d, yang2023deformable3dgs} proposed to use MLPs, 4D primitives and hexplane to represent the temporal information, achieving dynamic scene reconstruction. Despite the rendering performance, 3DGS utilizes explicit scene representation This novel technique has been rapidly applied in the surgical reconstruction task~\cite{liu2024endogaussian, huang2024endo4dgsendoscopicmonocularscene, zhu2024deformable} for high-quality deformable tissue reconstruction. However, 3DGS lacks a definition of surface normal, resulting in view-inconsistent geometry. To address this challenge, 2DGS~\cite{Huang2DGS2024} is presented with well-defined surface normals. In this work, we present a novel SLAM system with 2DGS as scene representation, providing more view-consistent reconstruction than 3DGS-based SLAM.

\subsection{Endoscopic SLAM}

Endoscopic SLAM faces unique challenges~\cite{fu2021future} due to constrained environments, limited field of view, scarce textures, deformation, \textit{etc}. Previous works~\cite{grasa2013visual, mahmoud2017slam, mahmoud2018live} adopt feature-based tracking to address the illumination variation. Several works~\cite{ma2021rnnslam, qiu2018endoscope, turan2018unsupervised, liu2020extremely} provide solutions such as markers tracking and correspondence analysis against the texture scarcity issue. To improve the quality of dense reconstruction, some approaches~\cite{gu2022vision, sageslam, ma2021rnnslam, posner2023c, rau2023bimodal, endodepth, afsfm, wei2022stereo} incorporate deep neural network for dense estimation and tracking. Graph-based feature~\cite{rodriguez2024nr} is applied for non-rigid tissue tracking. Most recently, EndoGSLAM~\cite{wang2024endogslam} facilitates endoscopic reconstruction with real-time novel view synthesis. The high efficiency allows surgeons to interact and inspect any region of the 3D scene. Despite these advances, endoscopic SLAM still remains challenging in terms of reconstruction quality and time efficiency. In this paper, we introduce a novel endoscopic SLAM that achieves state-of-the-art (SOTA) geometry accuracy for real-time reconstruction.

%% file: sections/3_system.tex
\section{Methodology}
\begin{figure*}[ht!]
    \centering
    \includegraphics[width=\textwidth]{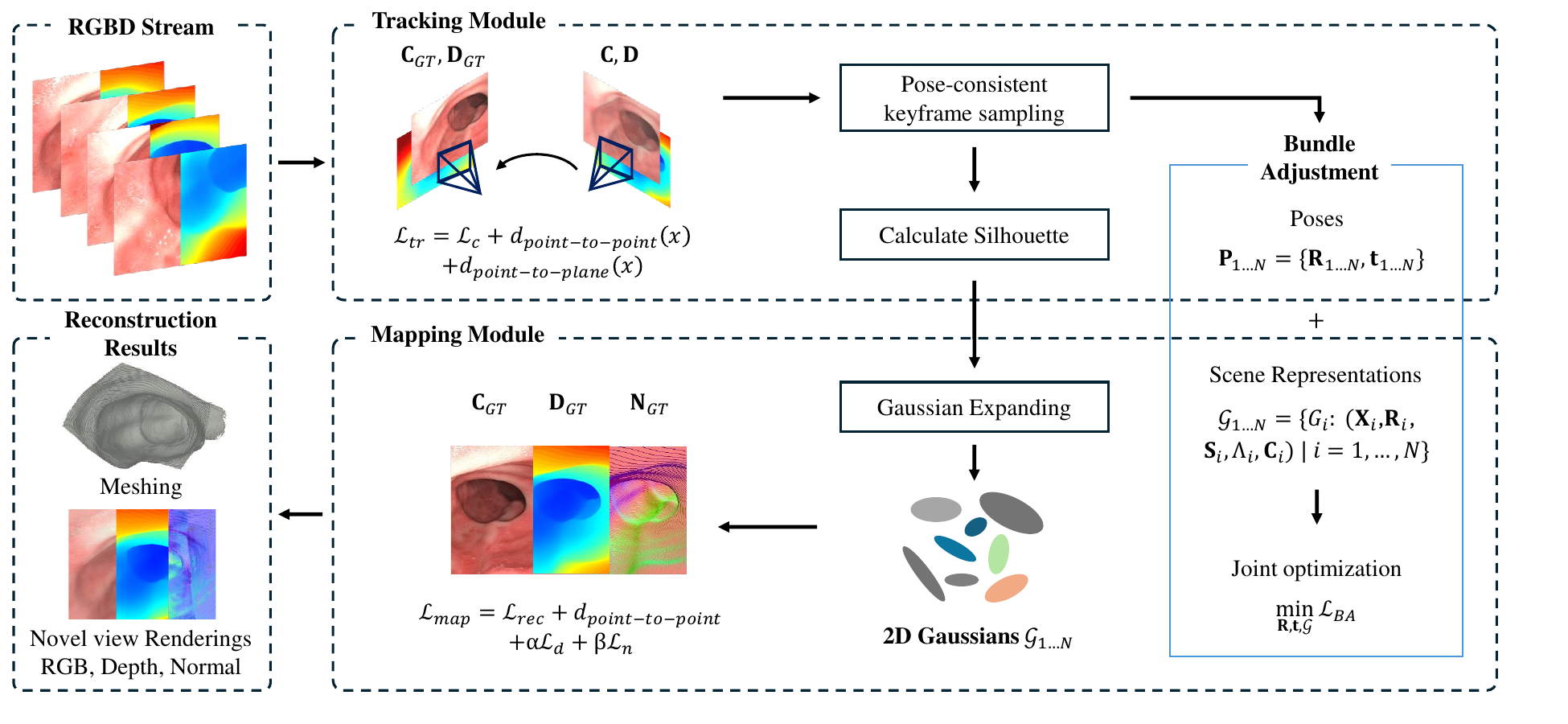}
    \caption{\textbf{Overview of Endo-2DTAM.} Our proposed system consists of three modules: the tracking module, the mapping module, and the bundle adjustment. The tracking module takes the incoming RGBD frame as input and tracks the camera pose. Then the frame is added to the candidate list for the pose-consistent keyframe selection. In the mapping module, we first expand 2D Gaussians with the new frame and then update 2D Gaussians with the selected keyframes. The selected keyframes are also used for bundle adjustment for joint optimization of poses and 2D Gaussians. 
    }
\label{fig2}
\end{figure*}
Endo-2DTAM is a dense RGB-D SLAM system based on 2D Gaussian Splatting. Fig.~\ref{fig2} shows the overview of our proposed system. In this section, we present the details of the system from the following aspects: 2D Gaussian representation~(\ref{2dgs}), tracking~(\ref{tracking}), gaussian expanding and keyframe sampling~(\ref{keyframe}), mapping~(\ref{mapping}), and bundle adjustment~(\ref{ba}).

\subsection{2D Gaussian Scene Representation} 
\label{2dgs}
To represent the endoscopic scene with accurate geometry and achieve real-time high-quality rendering, we model the endoscopic scene with a set of 2D Gaussians:
\begin{equation}
    \mathcal{G} = \{G_i: (\mathbf{X}_i, \mathbf{R}_i, \mathbf{S}_i, \Lambda_i, \mathbf{C}_i)| i=1,...,N\}
\end{equation}
We define each 2D Gaussian $G_i$ with the position $\mathbf{X}_i \in \mathbb{R}^3$, rotation $\mathbf{R}_i \in \mathbb{R}^{3\times3}$, scaling factor $\mathbf{S}_i \in \mathbb{R}^{3\times3}$, opacity $\Lambda_i \in \mathbb{R}$, and RGB color $\mathbf{C}_i \in \mathbb{R}^3$. The rotation matrix is defined by $\mathbf{R}_i=[\mathbf{t}_u,\mathbf{t}_v, \mathbf{t}_w]$, where $\mathbf{t}_u,\mathbf{t}_v$ are two orthogonal tangential vectors and $\mathbf{t}_w=\mathbf{t}_u\times\mathbf{t}_v$ is the primitive normal. Following~\cite{Huang2DGS2024}, we define the scaling matrix $\mathbf{S}_i$ as a diagonal matrix whose last entry is zero and the other two are $(s_u,s_v)$.

Given the optimized 2D Gaussian representation and a pose $\mathbf{P}$, we render the novel view with the rasterization method by~\cite{Huang2DGS2024}. The transformation between the UV space and world space can be represented by:
\begin{equation}
    \mathbf{H} = \begin{bmatrix}
        s_u \mathbf{t}_u & s_v \mathbf{t}_v & \boldsymbol{0} & \mathbf{X}_i \\
        0 & 0 & 0 & 1 \\
    \end{bmatrix} = \begin{bmatrix}
        \mathbf{R}_i\mathbf{S}_i & \mathbf{X}_i \\ 
        \boldsymbol{0} & 1\\
    \end{bmatrix}
\end{equation}

With the world to screen transformation $\mathbf{W}$ from $\mathbf{P}$, for a screen point ${\boldsymbol x}=(x, y)$ we have the ray $(xz, yz, z, 1)^\mathrm{T} = \mathbf{W} \mathbf{H} (u,v,1,1)^\mathrm{T}$, where $z$ is the depth for ray-splat intersection. In rasterization~\cite{Huang2DGS2024}, we inquire about the intersection point $\mathbf{u}({\boldsymbol x})=(u({\boldsymbol x}), v({\boldsymbol x}))$ in Gaussian's coordinate by:
\begin{equation}
    u({\boldsymbol x}) = \frac{\mathbf{h}_u^2 \mathbf{h}_v^4 - \mathbf{h}_u^4 \mathbf{h}_v^2}{\mathbf{h}_u^1 \mathbf{h}_v^2-\mathbf{h}_u^2 \mathbf{h}_v^1} \qquad v({\boldsymbol x}) = \frac{\mathbf{h}_u^4 \mathbf{h}_v^1 - \mathbf{h}_u^1 \mathbf{h}_v^4}{\mathbf{h}_u^1 \mathbf{h}_v^2-\mathbf{h}_u^2\mathbf{h}_v^1},
\label{eq:uv_intersection}
\end{equation}
where $\mathbf{h}_u = (\mathbf{W} \mathbf{H})^\mathrm{T} {\mathbf{h}_x}  \quad \mathbf{h}_v = (\mathbf{W} \mathbf{H})^\mathrm{T} {\mathbf{h}_y}$ are the two 4D homogeneous plane and $\mathbf{h}_u^i \mathbf{h}_v^i$ are the i-th parameter. The final color $\mathbf{c}({\boldsymbol x})$ is rendered as:
\begin{equation}
\mathbf{c}({\boldsymbol x}) = \sum_{i=1}T_i\mathbf{c}_i\,\alpha_i\,\hat{G}_i(\mathbf{u}({\boldsymbol x})),
\end{equation}
where $T_i= \prod_{j=1}^{i-1} (1 - \alpha_j\,\hat{G}_j(\mathbf{u}({\boldsymbol x})))$ is the visibility of the 2D Gaussian, $\mathbf{c}_i, \alpha_i$ are the corresponding RGB color and opacity, and $\hat{G}_i$ is its 2D Gaussian value after the object-space low-pass filtering as in~\cite{Huang2DGS2024}. 

The depth map of the given pose is rendered as:
\begin{equation}
    D({\boldsymbol x}) = \sum_{i}\omega_iz_i / ({\sum_{i}\omega_i + \epsilon}),
\end{equation}
where $\omega_i=T_i\alpha_i\hat{G}_i(\mathbf{u}({\boldsymbol x}))$ is the weight contribution of the $i$-th Gaussian, $z_i$ is the depth of the intersection point, $\epsilon$ is a small constant to prevent zero divisor. 

\subsection{Tracking}
\label{tracking}
Similar to previous work~\cite{keetha2024splatam}, we assume a constant velocity for the novel pose initialization. Given two previous poses $\mathbf{P}_{t-1}, \mathbf{P}_{t-2}$, we calculate the velocity as $\Delta(\mathbf{P}_{t-1}, \mathbf{P}_{t-2})$ and initialize $\mathbf{P}_t = \mathbf{P}_{t-1}+\Delta(\mathbf{P}_{t-1}, \mathbf{P}_{t-2})$. Our camera pose is then optimized iteratively by gradient descent. 

To perform geometrically robust tracking, we propose a surface normal aware tracking regularization with the projective point-to-point distance $d_{point-to-point}({\boldsymbol x})$, and point-to-plane distance $d_{point-to-plane}({\boldsymbol x})$. Let $\mathbf{x}$ be the real-world point projected from ${\boldsymbol x}$, $\mathbf{x}_{GT}$ is the point from ground truth depth. We define $d_{point-to-point}({\boldsymbol x})$ as:
\begin{equation}
\begin{aligned}
    d_{point-to-point}({\boldsymbol x}) &= ||z(\mathbf{x}_{GT})-z(\mathbf{x})||_1\\
    &= || \mathbf{D}_{GT}-\mathbf{D}||_1,
\end{aligned}
\end{equation}
where $z(\cdot)$ is the z value, $\mathbf{D}_{GT}, \mathbf{D}$ are ground truth depth and rendered depth.

Since tracking with only perspective point-to-point distance lacks surface information, we propose to use the normal aware point-to-plane distance:
\begin{equation}
    d_{point-to-plane}({\boldsymbol x}) = ||(\mathbf{x}_{GT} - \mathbf{x})^T\mathbf{N}_{GT}||_1
\end{equation}
\begin{equation}
    \mathbf{N}_{GT} = \frac{\nabla_x \mathbf{x}_{GT} \times \nabla_y \mathbf{x}_{GT}}{|\nabla_x \mathbf{x}_{GT} \times \nabla_y \mathbf{x}_{GT}|},
\end{equation}
where $\mathbf{N}_{GT}$ is the ground truth normal by taking the finite difference from the nearby points. $\nabla_x$ and $\nabla_y$ are the differences in the x-axis and y-axis, respectively. 

By considering the appearance loss, we formulate the final loss function for tracking as follows:
\begin{equation}
    \begin{aligned}
    \mathcal{L}_{tr} = \mathcal{L}_{c}+d_{point-to-point}({\boldsymbol x})+d_{point-to-plane}({\boldsymbol x}),
    \end{aligned}
\end{equation}
where $\mathcal{L}_{c}=||\hat{\mathbf{C}}({\boldsymbol x})-\mathbf{C}_{GT}({\boldsymbol x})||_1$ is the L1 loss of the render image, $\hat{\mathbf{C}}$ is the render color after the affine exposure adjustment~\cite{Matsuki:Murai:etal:CVPR2024},
and $\mathbf{C}_{GT}$ is the ground truth color.

\subsection{Gaussian Expanding and Keyframe Sampling}
\label{keyframe}
To maintain a stable optimization during the camera tracking and mapping process, we adopt the Gaussian expanding strategy~\cite{keetha2024splatam}. We add the incoming frame to the keyframe candidate list and update the 2D Gaussian scene for every $k$ frame to incorporate the unobserved tissues. To update the 2D Gaussians, we utilize the rendered silhouette map $
    S({\boldsymbol x}) = \sum_{i}\omega_i / ({\sum_{i}\omega_i + \epsilon})$ to represent the observed region.
We expand the 2D Gaussian representation with the area that has a smaller silhouette value ($S({\boldsymbol x}) < \rho_e$) than the threshold $\rho_e$ and the region with ground truth geometry above the reconstructed tissue surface. The new Gaussians are initialized with the incoming image and depth. 

After updating, we sample $n$ keyframes from the previous candidate list for further mapping and bundle adjustment. We propose to use a pose-consistent strategy considering both the translation and rotation errors.  Inspired by~\cite{wang2024endogslam}, we define the probability of each candidate as:
\begin{equation}
    P = \sum_{l\in \{ d_l, r_l, t_l\}}\log_2\left(1 + \frac{1}{l + s}\right),
\end{equation}
where $d_l$ is the L2 distance of positions, $r_t$ is the magnitude error of rotation quaternions and $t_l$ is the time-lapse with respect to the current frame. $s$ is the scaler constant set to $0.2$. The probability of each frame is normalized so that $\sum P_i = 1-P_c$, where $P_c$ is the pre-defined probability of the current frame. All candidate frames are sorted and sampled by the cumulative distribution function of the probability. 

\input{figures/result}

\subsection{Mapping}
\label{mapping}
Given the optimizer camera poses from the tracking module, we update the 2D Gaussian scene representation iteratively with the gradient descent method with the fixed camera poses. To optimize the 2D Gaussian representations with compact appearance and depth rendering, we follow~\cite {Huang2DGS2024} to define the depth distortion loss for minimizing the distance between the ray-splat intersections:
\begin{equation}
    \mathcal{L}_{d} = \sum_{i,j}\omega_i\omega_j|z_i-z_j|,
\end{equation}
where $w_i$, $z_i$ are the weight and depth for the $i$-th intersection. Inspired by~\cite{Huang2DGS2024}, we also constrain the surface normal of the 2D Gaussians with the normal consistency loss:
\begin{equation}
    \mathcal{L}_{n} = \sum_{i} \omega_i (1-\mathbf{n}_i^\mathrm{T}\mathbf{N}_{GT}),
\end{equation}
where $i$ are the indexes over the intersected splats along the ray. $\mathbf{n}$ is the normal of the splat with respect to the camera. 

We present the final mapping loss as follows:
\begin{equation}
\begin{aligned}
    \mathcal{L}_{map} = \mathcal{L}_{rec} + d_{point-to-point} +\alpha \mathcal{L}_{d} + \beta \mathcal{L}_{n},
\end{aligned}
\end{equation}
where $\mathcal{L}_{rec} = (1-\lambda)||\hat{\mathbf{C}}({\boldsymbol x})-\mathbf{C}_{GT}({\boldsymbol x})||_1 + \lambda \mathcal{L}_{D-SSIM}$ is the reconstruction loss for color. $\mathcal{L}_{D-SSIM}$ is the SSIM loss from~\cite{3dgs}. $\lambda$ is 0.2, $\alpha$ is $1000$, and $\beta$ is $0.05$ as in~\cite{Huang2DGS2024}.

\subsection{Bundle Adjustment}
\label{ba}

Given a series of camera poses, and the optimized 2D Gaussian scene representation after mapping, we use a gradient descent-based bundle adjustment (BA) for joint pose and scene optimization. We perform BA every $100$ tracking iteration by selecting keyframes from the keyframe candidate list for optimization. We use the keyframe selection strategy described in~\ref{keyframe} and we randomly arrange the selected keyframe to update the pose and 2D Gaussians by a total of $200$ iterations. Given the pose $\mathbf{P}=\{\mathbf{R}, \mathbf{t}\}$ and 2D Gaussians $\mathcal{G}$, the optimization is achieved by minimizing the loss $\mathcal{L}_{BA}$:
\begin{equation}
\begin{aligned}
    \min_{\mathbf{R}, \mathbf{t}, \mathcal{G}}\mathcal{L}_{BA} &= \mathcal{L}_c+d_{point-to-point}({\boldsymbol x})\\
    &+d_{point-to-plane}({\boldsymbol x})+\alpha \mathcal{L}_{d} + \beta \mathcal{L}_{n}
    \label{eq:track_loss}
\end{aligned}
\end{equation}

%% file: figures/result.tex
\begin{figure*}[ht!]
\centering
{\footnotesize
\setlength{\tabcolsep}{1pt}
\renewcommand{\arraystretch}{0.9}
\newcommand{\sz}{0.18}
\begin{tabular}{cccccc}
\rotatebox[origin=c]{90}{{\scriptsize EndoGSLAM~\cite{wang2024endogslam}}} & 
\raisebox{-0.5\height}{\includegraphics[width=\sz\linewidth]{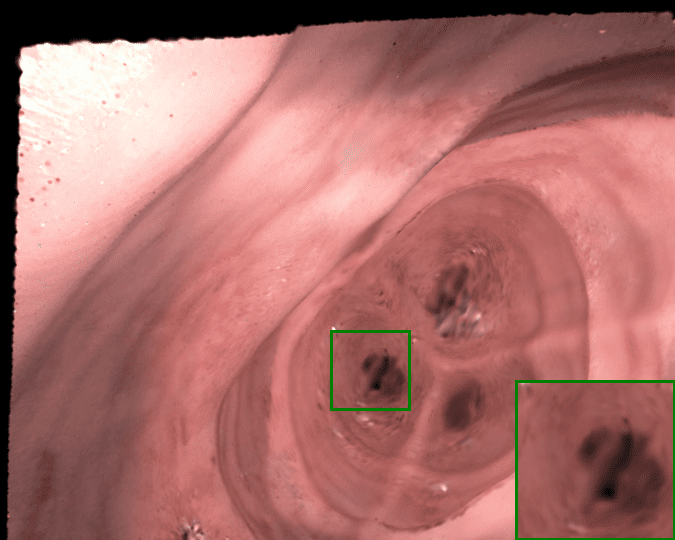}} & 
\raisebox{-0.5\height}{\includegraphics[width=\sz\linewidth]{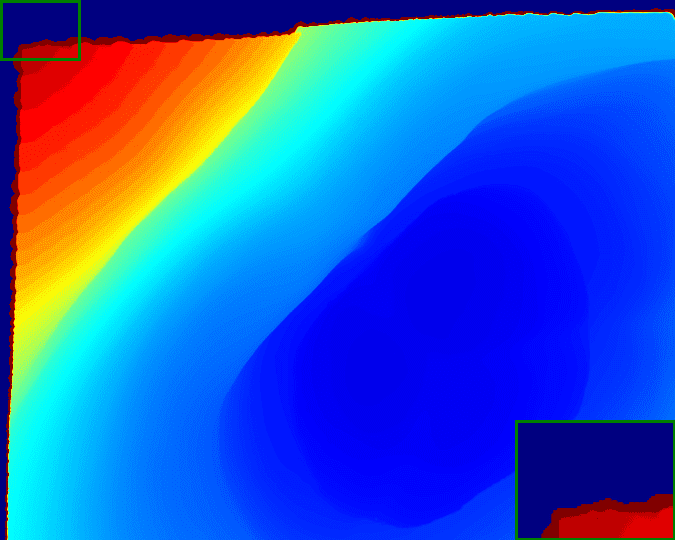}} &
\raisebox{-0.5\height}{\includegraphics[width=\sz\linewidth]{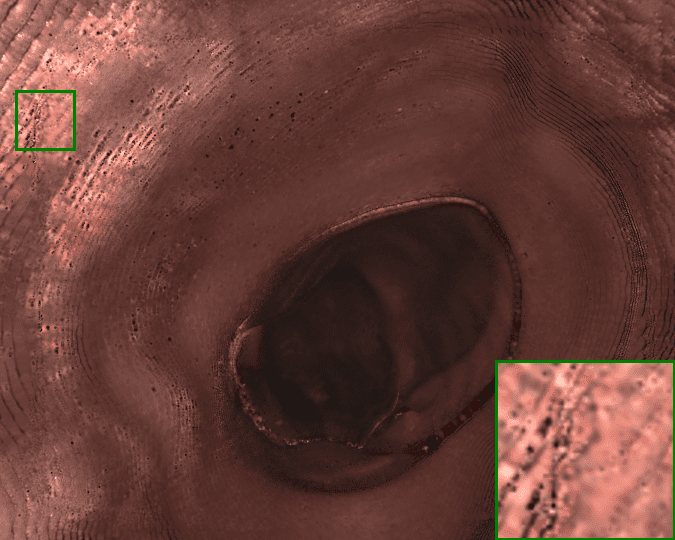}} &
\raisebox{-0.5\height}{\includegraphics[width=\sz\linewidth]{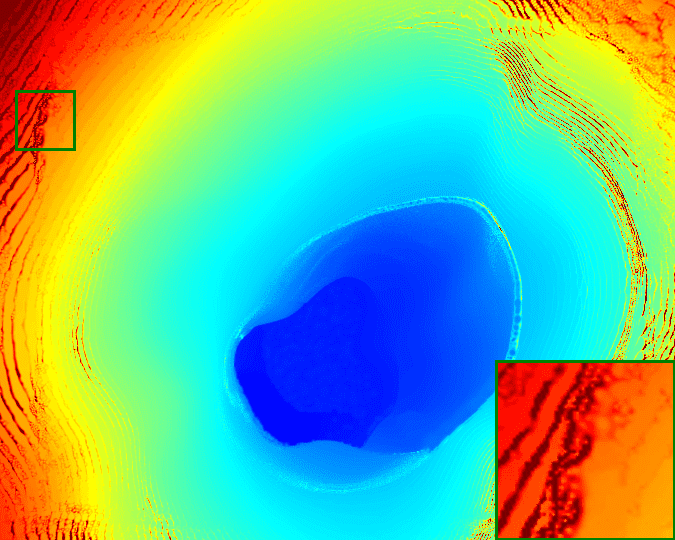}} & 
\raisebox{-0.5\height}{\includegraphics[width=\sz\linewidth]{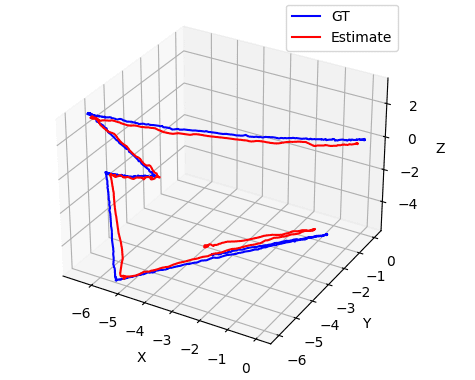}}
\\
\rotatebox[origin=c]{90}{\textbf{\scriptsize Ours}} & 
\raisebox{-0.5\height}{\includegraphics[width=\sz\linewidth]{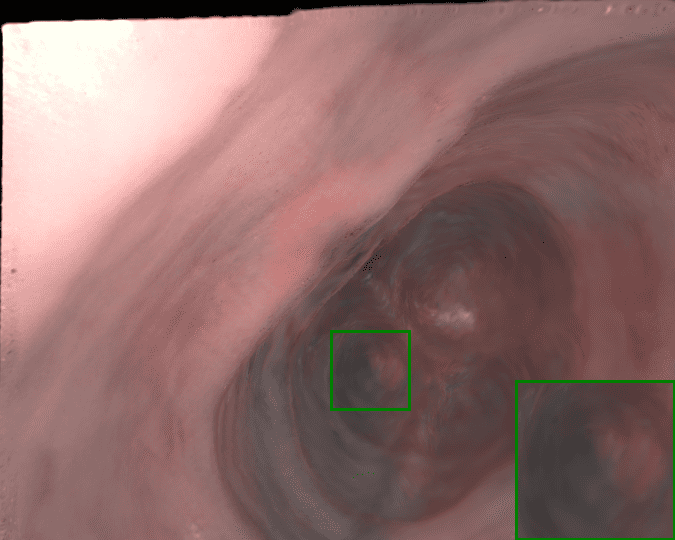}} & 
\raisebox{-0.5\height}{\includegraphics[width=\sz\linewidth]{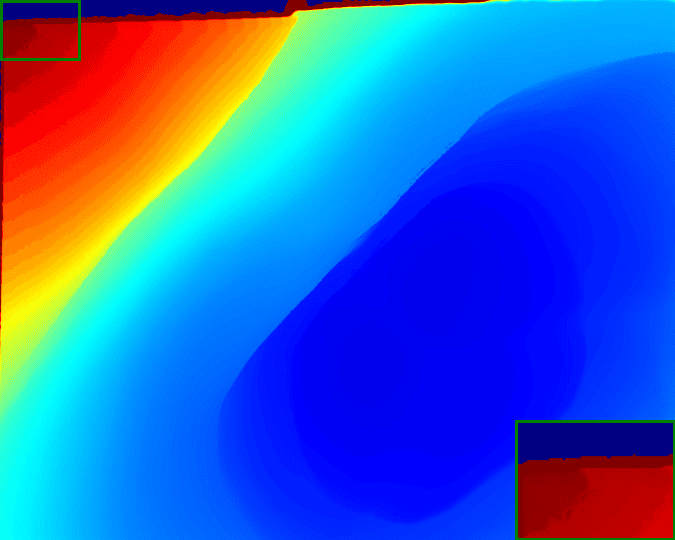}} &
\raisebox{-0.5\height}{\includegraphics[width=\sz\linewidth]{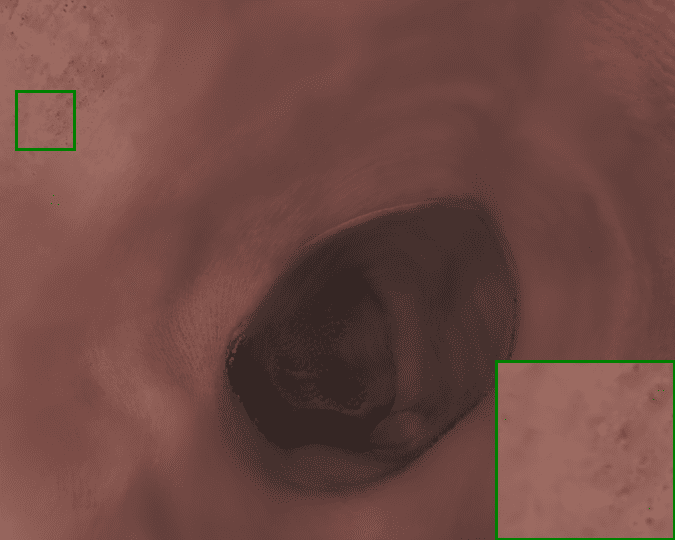}} &
\raisebox{-0.5\height}{\includegraphics[width=\sz\linewidth]{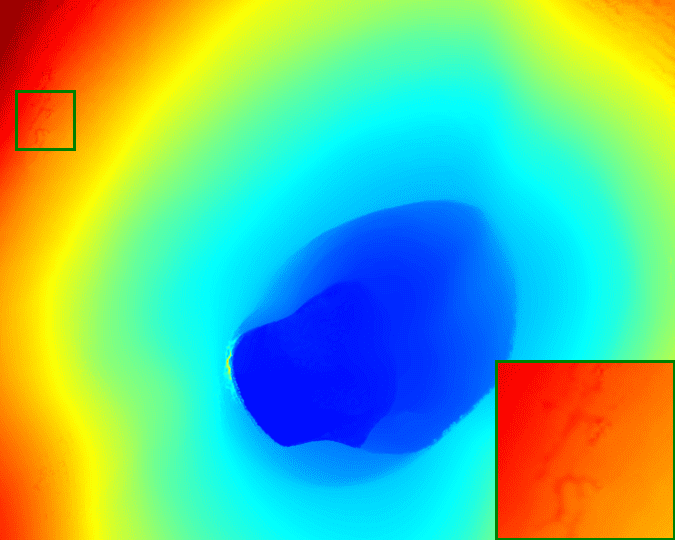}} & 
\raisebox{-0.5\height}{\includegraphics[width=\sz\linewidth]{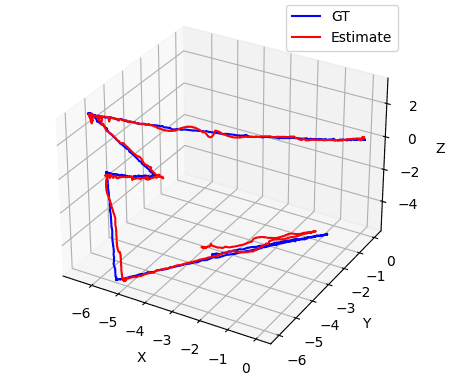}}
\\
\rotatebox[origin=c]{90}{\scriptsize GT} & 
\raisebox{-0.5\height}{\includegraphics[width=\sz\linewidth]{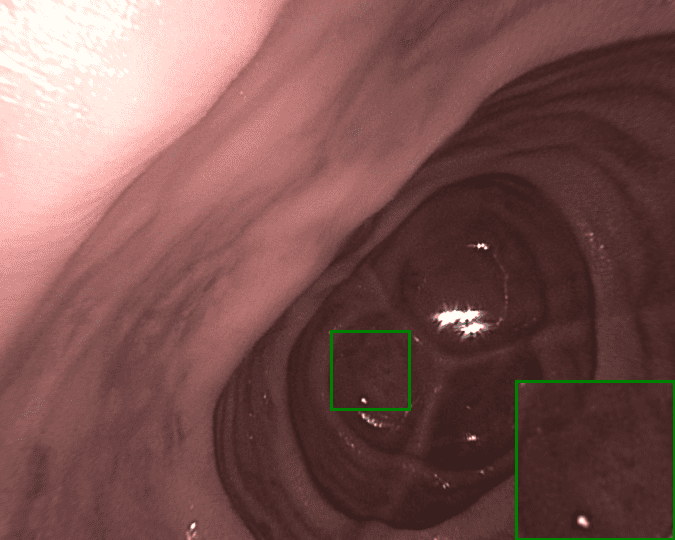}} & 
\raisebox{-0.5\height}{\includegraphics[width=\sz\linewidth]{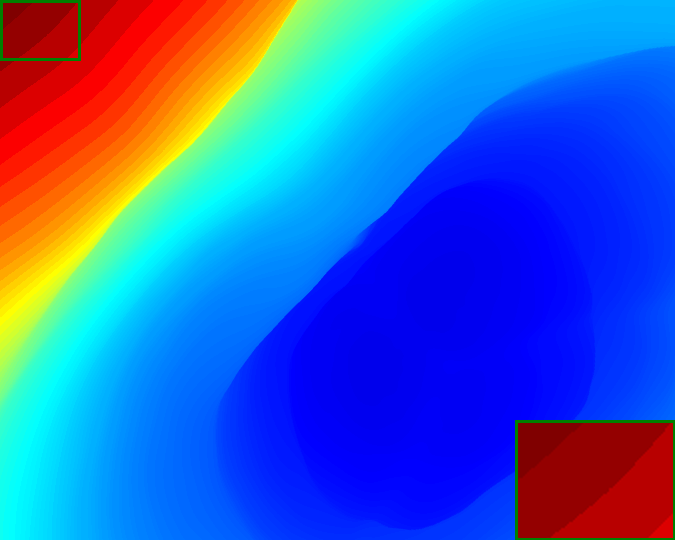}} &
\raisebox{-0.5\height}{\includegraphics[width=\sz\linewidth]{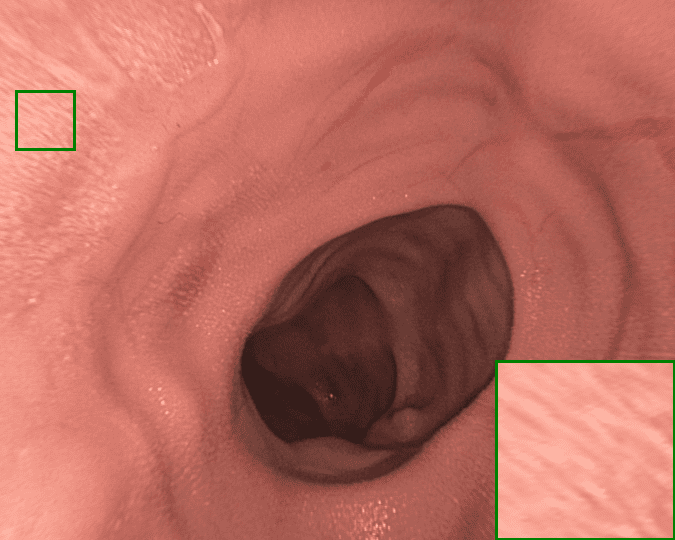}} &
\raisebox{-0.5\height}{\includegraphics[width=\sz\linewidth]{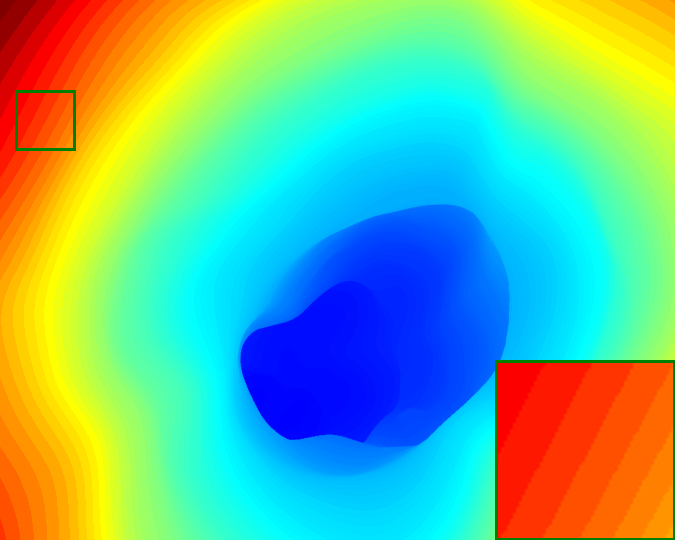}} & 
\raisebox{-0.5\height}{\includegraphics[width=\sz\linewidth]{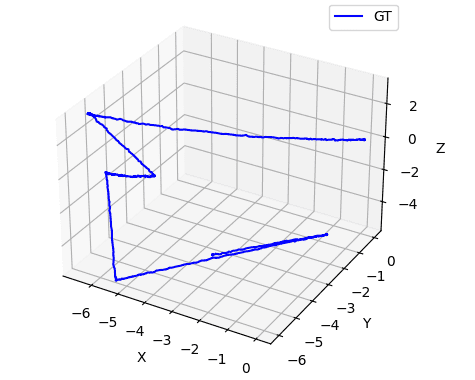}}
\\
& \multicolumn{2}{c}{\textit{cecum\_t2\_b}} & \multicolumn{2}{c}{\textit{sigmoid\_t2\_a}} & \multicolumn{1}{c}{\textit{cecum\_t3\_a}}
\\
\end{tabular}
}
\caption{\textbf{Qualitative Result on C3VD~\cite{bobrow2023}.} We compare our method with the SOTA EndoGSLAM~\cite{wang2024endogslam} for dense endoscopic SLAM. Our method generates more robust color and depth reconstruction as shown by results from \textit{cecum\_t2\_b}, \textit{sigmoid\_t2\_a}. Our method also estimates a more precise trajectory demonstrated by results from \textit{cecum\_t3\_a}.
}
\label{fig: qualitative}
\end{figure*}

%% file: sections/4_experiment.tex
\section{Experiment}

\begin{table*}[ht!]
    \setlength\tabcolsep{5pt}
    \centering
    \caption{Quantitative results on the C3VD dataset.}
    \resizebox{0.85\linewidth}{!}{{\input{tables/metrics}}} 
    \label{tab: metrics}
\end{table*}

In the experiment section, we demonstrate the evaluation result of our proposed Endo2DTAM on the public dataset for endoscopic SLAM across various intrabody surgery scenes. We compare our method with the state-of-the-art SLAM methods, which show our superior performance in endoscopic scene reconstruction.

\subsection{Experiment Setup}

\subsubsection{Dataset} To evaluate, we conduct experiments on the Colonoscopy 3D Video Dataset (C3VD)~\cite{bobrow2023}. We select 10 video clips of $675 \times 540$ resolution following~\cite {wang2024endogslam}, each including an average of 638 frames. 

\subsubsection{Baselines} We benchmark our method by comparing with the traditional method ORB-SLAM3~\cite{orbslam}, the implicit neural NICE-SLAM~\cite{niceslam}, the endoscopic SLAM Endo-Depth~\cite{endodepth}, and the state-of-the-art 3DGS-based SLAM EndoGSLAM~\cite{wang2024endogslam} for endoscopy. For a fair comparison, all baselines are provided with RGB-D input. 

\subsubsection{Metrics} To evaluate the camera trajectory, we report the absolute trajectory (ATE, mm). For depth, we use Root Mean Square Error (RMSE, mm). We also report standard photometric rendering quality metrics (PSNR, SSIM, and LPIPS) for novel view synthesis.

\subsubsection{Implementation Details} We run all experiments under the settings of Core 13700K CPU, RTX 4090 GPU, and Ubuntu 20.04. Our SLAM system is implemented into three versions: \textit{Endo-2DTAM-Base}, \textit{Endo-2DTAM-Small}, and \textit{Endo-2DTAM-Tiny}. The Endo-2DTAM-Base version utilizes $p_c = 0.1$, the keyframe candidate list is updated for every 8 frames. Both tracking and mapping have 15 iterations per frame. In Endo-2DTAM-Small, tracking has 10 iterations per frame, mapping has 10 iterations per 2 frames, and we use $1/2$ resolution. In Endo-2DTAM-Tiny, tracking has 8 iterations per frame, mapping has 8 iterations per 2 frames, and we use $1/4$ resolution. For both Endo-2DTAM-Small and Endo-2DTAM-Tiny, $p_c$ is set to $0.5$, and the keyframe candidate list is updated for every 4 frames. We use $\rho_e = 0.5$ for all the three versions.

\subsection{Experiment Results}
 We evaluate three versions of our method on the C3VD dataset as shown in TABLE~\ref{tab: metrics}. Following~\cite{wang2024endogslam}, we split each scene into training and testing sets. We compare the performance of Endo-2DTAM with other baselines in the following aspects: the appearance, geometry of the novel view renderings from the testing set, and the camera trajectory of all frames in the training set. For appearance reconstruction, our method achieves the highest SSIM of $0.77 \pm 0.07$, indicating that the renderings by our method are closest to human perception. In terms of depth reconstruction, Endo-2DTAM-Base significantly outperforms all other baselines with the lowest $1.87 \pm 0.63$ RMSE. While maintaining accurate mapping, our method also achieves a competitive result on trajectory estimation by the ATE of $0.33 \pm 0.22$. Qualitative results in Fig.~\ref{fig: qualitative} also show that our method outperforms the 3DGS-based method with more robust renderings and accurate trajectory estimation. Different from the large performance gap from EndoGSLAM-H to EndoGSLAM-R, our Endo-2DTAM-Small and Endo-2DTAM-Tiny versions only have depth error increases of $19.7\%$ and $27.8\%$ with respect to the Endo-2DTAM-Base model, while the trajectory errors of our Small and Tiny versions also have small variances of $15.5\%$ and $21.2\%$. These results suggest that our method provides accurate geometry and robustness across various resolutions for surgical reconstruction, securing further surgical robotics tasks such as planning and navigating.

\begin{table}[t]
    \setlength\tabcolsep{5pt}
    \centering
    \caption{Time Performance on the C3VD dataset.}
    \resizebox{\linewidth}{!}{{\input{tables/times}}} 
    \label{tab: time}
\end{table}

\begin{table}[t]
    \setlength\tabcolsep{5pt}
    \centering
    \caption{Tracking Loss Ablation on C3VD/\textit{trans$\_$t1$\_$b}.}
    \resizebox{0.95\linewidth}{!}{{\input{tables/ablation_track}}} 
    \label{ab_track}
\end{table}

\subsection{Performance Analysis}
In TABLE~\ref{tab: time}, we evaluate our runtime from the following three aspects: the tracking and mapping time per frame, and the online rendering speed of the novel view. Although ORB-SLAM3~\cite{orbslam} and Endo-Depth~\cite{endodepth} achieve faster tracking and mapping for processing one frame, they rely on post-processing volumetric fusion for dense scene reconstruction and are unable to perform online rendering. NICE-SLAM~\cite{niceslam} achieves high geometric accuracy but lacks efficiency in real-time tracking and mapping. Compared with the SOTA 3DGS-based EndoGSLAM~\cite{wang2024endogslam}, Endo-2DTAM-Base maintains competitive tracking efficiency and uses less time for mapping. Additionally, our Small and Tiny versions have a better trade-off than EndoGSLAM-R with runtimes of 5.53 fps, and 9.82 fps, and even achieve higher tracking and reconstruction accuracy.

\subsection{Ablation Study}

We also evaluate Endo-2DTAM with ablation experiments on the tracking losses, the mapping modality used for supervision, and the keyframe sampling strategy. Results in TABLE~\ref{ab_track} show that our surface normal-aware tracking with both point-to-point, point-to-plane distance, and color loss combination effectively stabilizes the trajectory estimation and enhances reconstruction quality. In TABLE~\ref{ab_map}, we validate the optimal choice of supervision modalities that the utilization of color, depth, and surface normal achieves the best performance. In Fig.~\ref{fig: ab_normal}, we also demonstrate the qualitative result of rendered surface normal with different choices of mapping modalities for supervision. As demonstrated in TABLE~\ref{ab_keyframe}, our pose-consistent keyframe sampling approach surpasses other strategies, achieving the most significant enhancement. We further validate that incorporating bundle adjustment results in an overall improvement.

\input{figures/ablation_rendering}
\begin{table}[t!]
    \setlength\tabcolsep{5pt}
    \centering
    \caption{Mapping Modality Ablation on C3VD/\textit{trans$\_$t1$\_$b}.}
    \resizebox{0.9\linewidth}{!}{{\input{tables/ablation_map}}} 
    \label{ab_map}
\end{table}

\begin{table}[t!]
    \setlength\tabcolsep{5pt}
    \centering
    \caption{Sampling Strategy and BA Ablation on C3VD/\textit{trans$\_$t1$\_$b}.}
    \resizebox{0.95\linewidth}{!}{{\input{tables/ablation_keyframe}}} 
    \label{ab_keyframe}
\end{table}

%% file: tables/metrics.tex
\begin{tabular}{c|ccccc}
\toprule
Methods  & PSNR$\uparrow $        & SSIM$\uparrow$         & LPIPS$\downarrow$         & RMSE(mm)$\downarrow$            & ATE (mm)$\downarrow$    \\ \midrule
ORB-SLAM3\cite{orbslam}   & 17.89 { $\pm$} 2.31   &  0.64 { $\pm$} 0.10  &  0.35 { $\pm$} 0.06  &  7.72 { $\pm$} 2.65  &  \textbf{0.32} { $\pm$} \textbf{0.09} \\
NICE-SLAM\cite{niceslam} & 22.07 { $\pm$} 4.12   & 0.73 { $\pm$} 0.13  & 0.33 { $\pm$} 0.07  & 1.88 { $\pm$} 1.04   & 0.48 { $\pm$} 0.33  \\
Endo-Depth\cite{endodepth} & 18.13 { $\pm$} 2.43   & 0.64 { $\pm$} 0.09 & 0.33 { $\pm$} 0.06  & 5.10 { $\pm$} 2.39   &  1.25 { $\pm$} 0.98 \\
EndoGSLAM-H~\cite{wang2024endogslam} & \textbf{22.16} { $\pm$} \textbf{2.66}   & 0.77 { $\pm$} 0.08  & \textbf{0.22} { $\pm$} \textbf{0.05}    &  2.17 { $\pm$} 1.26     & 0.34 { $\pm$} 0.21    \\
EndoGSLAM-R~\cite{wang2024endogslam} & 18.37 { $\pm$} 2.17   & 0.67 { $\pm$} 0.10  & 0.30 { $\pm$} 0.07    & 4.33 { $\pm$} 2.39    &  1.23 { $\pm$} 0.90 
      \\ \midrule
Endo-2DTAM-Base & 20.21 { $\pm$} 1.76   & \textbf{0.77} { $\pm$} \textbf{0.07}  & 0.30 { $\pm$} 0.05    & \textbf{1.87} { $\pm$} \textbf{0.63}    &  0.33 { $\pm$} 0.22            \\
Endo-2DTAM-Small & 19.52 { $\pm$} 2.18   & 0.75 { $\pm$} 0.09  & 0.28 { $\pm$} 0.08    & 2.24 { $\pm$} 0.61    &  0.38 { $\pm$} 0.25            \\
Endo-2DTAM-Tiny & 19.28 { $\pm$} 2.48   & 0.76 { $\pm$} 0.09  & 0.24 { $\pm$} 0.09    & 2.39 { $\pm$} 0.62    &  0.40 { $\pm$} 0.26            \\
 \bottomrule
\end{tabular}

%% file: tables/times.tex
\begin{tabular}{c|ccc}
\toprule
\makecell{Methods} & \makecell{Tracking\\time/frame} & \makecell{Mapping\\time/frame} & \makecell{Rendering\\speed}  \\ \midrule
ORB-SLAM3~\cite{orbslam}   & \textbf{8.5ms}   & \textbf{32.3ms}  & $\times$ \\
NICE-SLAM~\cite{niceslam} & 140.29ms  & 2558.0ms   &  0.27 fps \\
Endo-Depth~\cite{endodepth} & 194.52ms   & 93.7ms  & $\times$  \\
EndoGSLAM-H~\cite{wang2024endogslam} & 151.4ms   & 268.0ms     & \textbf{100+ fps}   \\
EndoGSLAM-R~\cite{wang2024endogslam} & 62.4ms   & 65.1ms    & \textbf{100+ fps}   \\ \midrule
Endo-2DTAM-Base & 174.9ms   & 170.6ms     & \textbf{100+ fps}   \\
Endo-2DTAM-Small & 76.4ms   & 94.2ms    & \textbf{100+ fps}   \\
Endo-2DTAM-Tiny & 45.9ms   & 65.9ms    & \textbf{100+ fps}   \\
\bottomrule
\end{tabular}

%% file: tables/ablation_track.tex
\begin{tabular}{ccc|ccc}
\toprule
\makecell{$\mathcal{L}_c$} & \makecell{$d_{\substack{point-\\to-point}}$} & \makecell{$d_{\substack{point-\\to-plane}}$} & \makecell{PSNR$\uparrow$} & \makecell{Dep. L1\\ (mm)$\downarrow$} & \makecell{ATE\\ (mm)$\downarrow$} \\

\midrule
$\checkmark$ & $\times$ & $\times$ & 14.05 & 8.05 & 2.11\\
$\times$ & $\checkmark$ & $\times$ & 22.68 & 1.63 & 0.25 \\
$\times$ & $\times$ & $\checkmark$ & 22.62 & 1.45 & 0.26 \\
$\checkmark$ & $\checkmark$ & $\times$ & 22.85 & 1.44 & 0.24 \\
$\checkmark$ & $\times$ & $\checkmark$ & 22.78 & 1.52 & 0.29 \\
$\times$ & $\checkmark$ & $\checkmark$ & 22.74 & 1.46 & 0.24 \\
$\checkmark$ & $\checkmark$ & $\checkmark$ & \textbf{22.87} & \textbf{1.43} & \textbf{0.22} \\

\bottomrule
\end{tabular}

%% file: figures/ablation_rendering.tex
\begin{figure}[t!]
\centering
{\footnotesize
\setlength{\tabcolsep}{1pt}
\renewcommand{\arraystretch}{0.9}
\newcommand{\sz}{0.3}
\begin{tabular}{cccc}
\rotatebox[origin=c]{90}{{\scriptsize GT}} & 
\raisebox{-0.5\height}{\includegraphics[width=\sz\linewidth]{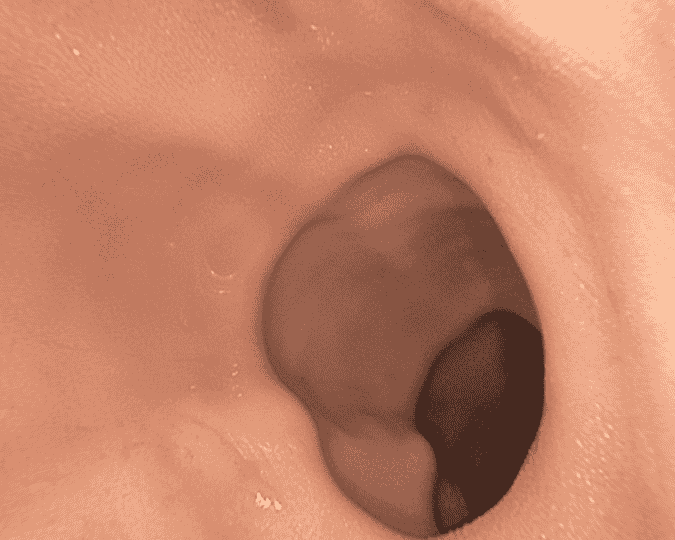}} & 
\raisebox{-0.5\height}{\includegraphics[width=\sz\linewidth]{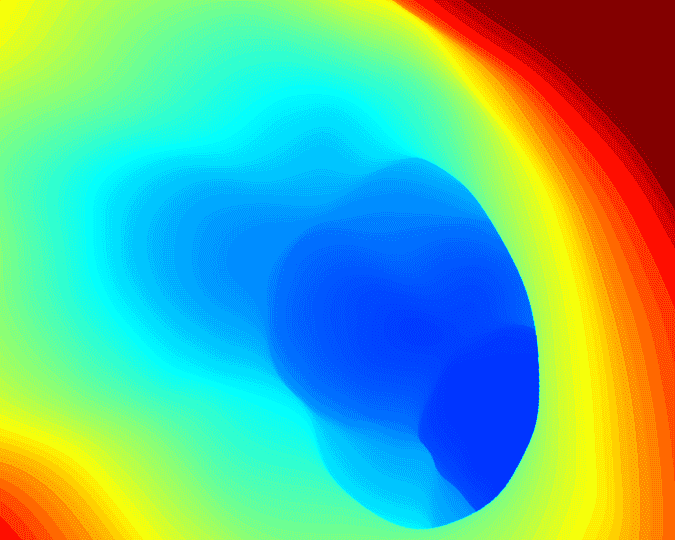}} &
\raisebox{-0.5\height}{\includegraphics[width=\sz\linewidth]{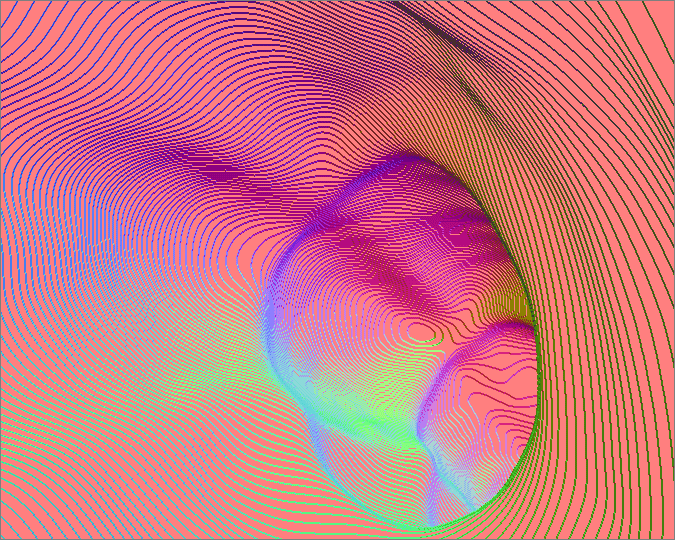}}
\\
& \multicolumn{1}{c}{GT Color} & \multicolumn{1}{c}{GT Depth} & \multicolumn{1}{c}{GT Normal}
\\
\rotatebox[origin=c]{90}{\scriptsize Rendered Normal} & 
\raisebox{-0.5\height}{\includegraphics[width=\sz\linewidth]{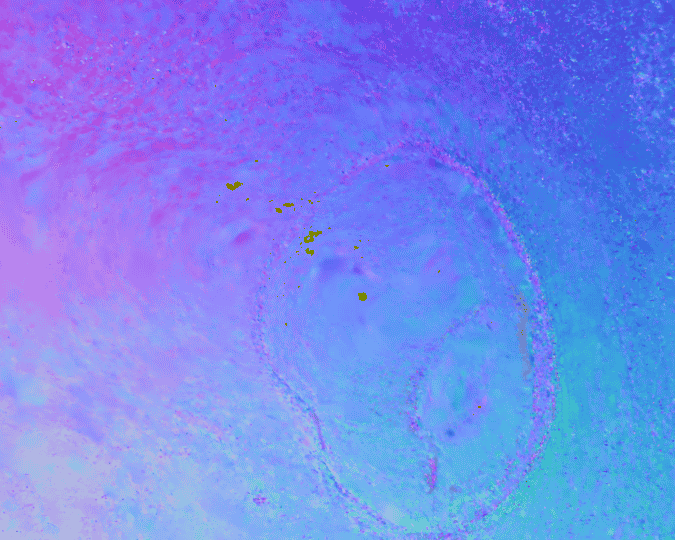}} & 
\raisebox{-0.5\height}{\includegraphics[width=\sz\linewidth]{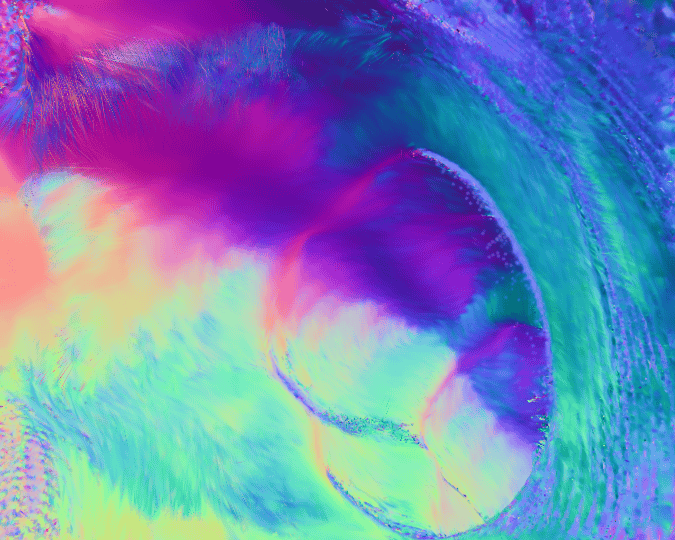}} &
\raisebox{-0.5\height}{\includegraphics[width=\sz\linewidth]{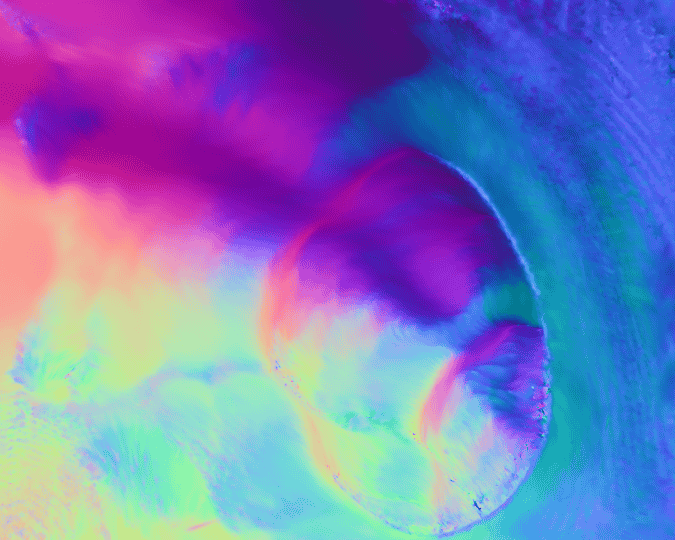}}
\\
& \multicolumn{1}{c}{\makecell{Only Color\\supervision}} & \multicolumn{1}{c}{\makecell{Color+Depth\\supervisions}} & \multicolumn{1}{c}{\makecell{Color+Depth+\\Normal supervisions}}
\\
\end{tabular}
}
\caption{\textbf{Surface Normal Comparison of Mapping Ablation.} We compare the rendered surface normal with different modalities as supervisions. Results demonstrate that the combination of all supervision with color, depth, and normal achieves the best quality.
}
\label{fig: ab_normal}
\end{figure}

%% file: tables/ablation_map.tex
\begin{tabular}{ccc|ccc}
\toprule
\makecell{Color} & \makecell{Depth} & \makecell{Surface\\Normal} & \makecell{PSNR$\uparrow$} & \makecell{Dep. L1\\ (mm)$\downarrow$} & \makecell{ATE\\ (mm)$\downarrow$} \\

\midrule
$\checkmark$ & $\times$ & $\times$ & 21.66 & 2.43 & 0.23\\
$\times$ & $\checkmark$ & $\times$ & $\times$ & 1.52 & 0.26 \\
$\times$ & $\times$ & $\checkmark$ & $\times$ & 2.85 & 0.30 \\
$\checkmark$ & $\checkmark$ & $\times$ & 21.83 & 1.50 & 0.27 \\
$\checkmark$ & $\times$ & $\checkmark$ & 21.69 & 2.07 & 0.24 \\
$\times$ & $\checkmark$ & $\checkmark$ & $\times$& 1.48 & 0.25 \\
$\checkmark$ & $\checkmark$ & $\checkmark$ & \textbf{22.87} & \textbf{1.43} & \textbf{0.22} \\

\bottomrule
\end{tabular}

%% file: tables/ablation_keyframe.tex
\begin{tabular}{c|ccc}
\toprule
\makecell{Sampling Strategy} & \makecell{PSNR$\uparrow$} & \makecell{Dep. L1\\ (mm)$\downarrow$} & \makecell{ATE\\ (mm)$\downarrow$} \\ \midrule
Overlap-based~\cite{keetha2024splatam} with BA   & 22.85  & 1.55  & 0.24 \\
Distance-based~\cite{wang2024endogslam} with BA & 22.79  & 1.52  & 0.23 \\
Ours with BA & \textbf{22.87} & \textbf{1.43} & \textbf{0.22}  \\
\midrule
Ours without BA & 19.24 & 1.46 & 0.25  \\
\bottomrule

\end{tabular}

%% file: sections/5_conclusions.tex
\section{CONCLUSIONS}

In this work, we introduce Endo-2DTAM, a novel dense SLAM system that utilizes 2D Gaussian Splatting to advance endoscopic reconstruction. Our method effectively mitigates the issue of view inconsistency in 3DGS-based SLAM, delivering surgeons with enhanced novel view synthesis, consistent depth estimation, and accurate surface normals essential for precise intraoperative visualization. Through extensive experiments, we show that Endo-2DTAM achieves state-of-the-art performance on the geometric reconstruction of surgical scenes while maintaining computationally efficient tracking. However, our current implementation may face challenges in scenes with rapid tissue deformations, which remain area for improvement. In future work, we will focus on reducing the depth dependence of Endo-2DTAM and integrate it into robotic-assisted surgical procedures for planning and navigation.
